\documentclass{article}

\usepackage{microtype}
\usepackage{graphicx}
\usepackage{subfigure}
\usepackage{booktabs} 

\usepackage{hyperref}
\usepackage[dvipsnames]{xcolor}

\usepackage[accepted]{icml2024}

\usepackage{amsmath}
\usepackage{amssymb}
\usepackage{mathtools}
\usepackage{amsthm}

\usepackage{amsmath,amsfonts,bm}

\def\Figref#1{Figure~\ref{#1}}

\def\Secref#1{Section~\ref{#1}}
\def\Appref#1{Appendix~\ref{#1}}

\def\eqref#1{equation~\ref{#1}}
\def\Eqref#1{Eq.~\ref{#1}}

\def\Tableref#1{Table~\ref{#1}}

\def\1{\bm{1}}

\newcommand{\normal}{\mathcal{N}}
\newcommand{\bld}[1]{\boldsymbol{#1}}

\def\rva{{\mathbf{a}}}

\def\rvf{{\mathbf{f}}}
\def\rvg{{\mathbf{g}}}
\def\rvh{{\mathbf{h}}}

\def\rvm{{\mathbf{m}}}

\def\rvw{{\mathbf{w}}}
\def\rvx{{\mathbf{x}}}
\def\rvy{{\mathbf{y}}}

\def\rmA{{\mathbf{A}}}
\def\rmB{{\mathbf{B}}}

\def\rmH{{\mathbf{H}}}

\def\rmJ{{\mathbf{J}}}

\def\rmM{{\mathbf{M}}}

\def\rmS{{\mathbf{S}}}

\def\rmX{{\mathbf{X}}}

\DeclareMathAlphabet{\mathsfit}{\encodingdefault}{\sfdefault}{m}{sl}
\SetMathAlphabet{\mathsfit}{bold}{\encodingdefault}{\sfdefault}{bx}{n}

\def\gB{{\mathcal{B}}}

\def\gD{{\mathcal{D}}}

\def\sR{{\mathbb{R}}}

\newcommand{\E}{\mathbb{E}}

\DeclareMathOperator*{\argmax}{arg\,max}
\DeclareMathOperator*{\argmin}{arg\,min}

\def\rvtheta{\bld{\mathbf{\theta}}}
\newcommand{\rmSigma}{\bld{\mathbf{\Sigma}}}
\newcommand{\rvsigma}{\bld{\mathbf{\sigma}}}

\newcommand{\rvmu}{\bld{\mathbf{\mu}}}
\newcommand{\rmLambda}{\mathbf{\Lambda}}
\newcommand{\rvalpha}{\bld{\mathbf{\alpha}}}
\newcommand{\rvlambda}{\bld{\mathbf{\lambda}}}

\newcommand{\MN}{BayesAgg-MTL }

\newcommand\ignore[1]{{}}

\usepackage[capitalize,noabbrev]{cleveref}

\theoremstyle{plain}

\theoremstyle{definition}

\theoremstyle{remark}

\usepackage[textsize=tiny]{todonotes}
\usepackage[normalem]{ulem}

\icmltitlerunning{Bayesian Uncertainty for Gradient Aggregation in Multi-Task Learning}

\begin{document}

\twocolumn[
\icmltitle{Bayesian Uncertainty for Gradient Aggregation in Multi-Task Learning}



\icmlsetsymbol{equal}{*}

\begin{icmlauthorlist}
\icmlauthor{Idan Achituve}{sch1,comp}
\icmlauthor{Idit Diamant}{comp}
\icmlauthor{Arnon Netzer}{comp}
\icmlauthor{Gal Chechik}{sch2}
\icmlauthor{Ethan Fetaya}{sch1}
\end{icmlauthorlist}

\icmlaffiliation{sch1}{Faculty of Engineering, Bar-Ilan University, Israel}
\icmlaffiliation{sch2}{Department of Computer Science, Bar-Ilan University, Israel}
\icmlaffiliation{comp}{Sony Semiconductor Israel}

\icmlcorrespondingauthor{Idan Achituve}{Idan.Achituve@Sony.com}

\icmlkeywords{Machine Learning, ICML}

\vskip 0.3in
]



\printAffiliationsAndNotice{}  

\begin{abstract}
As machine learning becomes more prominent there is a growing demand to perform several inference tasks in parallel. Running a dedicated model for each task is computationally expensive and therefore there is a great interest in multi-task learning (MTL). MTL aims at learning a single model that solves several tasks efficiently. Optimizing MTL models is often achieved by first computing a \textit{single} gradient per task and then aggregating the gradients for obtaining a combined update direction. However, this approach do not consider an important aspect, the sensitivity in the gradient dimensions. Here, we introduce a novel gradient aggregation approach using Bayesian inference. We place a probability distribution over the task-specific parameters, which in turn induce a \textit{distribution} over the gradients of the tasks. This additional valuable information allows us to quantify the uncertainty in each of the gradients dimensions, which can then be factored in when aggregating them. We empirically demonstrate the benefits of our approach in a variety of datasets, achieving state-of-the-art performance.
\end{abstract}

\section{Introduction}
\label{sec:intro}
In many application domains, there is a need to perform several machine learning inference tasks simultaneously. For instance, an autonomous vehicle needs to identify and detect objects in its vicinity, perform lane detection, track the movements of other vehicles over time, and predict free space around it, all in parallel and in real-time. In deep Multi-Task Learning (MTL) the goal is to train a single neural network (NN) to solve several tasks simultaneously, thus avoiding the need to have one dedicated model for each task \cite{caruana1997multitask}. Besides reducing the computational demand at test time, MTL also has the potential to improve generalization \cite{baxter2000model}. It is therefore not surprising that applications of MTL are taking central roles in various domains, such as vision \cite{achituve2021self, shamshad2023transformers, zheng2023deep}, natural language processing \cite{liu2019multi, zhou2023comprehensive}, speech \cite{michelsanti2021overview}, robotics \cite{devin2017learning, shu2018hierarchical}, and general scientific problems \cite{wu2018moleculenet} to name a few. 

However, optimizing multiple tasks simultaneously is a challenging problem that may lead to degradation in performance compared to learning them individually \cite{standley2020tasks, yu2020gradient}. To address this issue, one basic formula that many MTL optimization algorithms follow is to first calculate the gradient of each task's loss, and then aggregate these gradients according to some specified scheme. For example, several studies focus on reducing conflicts between the gradients before averaging them \cite{yu2020gradient, wang2020gradient}, others find a convex combination with minimal norm \cite{sener2018multi, desideri2012multiple}, and some use a game theoretical approach \cite{navon2022multi}. However, by relying only on the gradient these methods miss an important aspect, the sensitivity of the gradient in each dimension.



Our approach builds on the following observation - for each task, there may be many ``good" parameter configurations. Standard MTL optimization methods take only a single value into account, and as such lose information in the aggregation step. Hence, tracking all of the parameter configurations will yield a richer description of the gradient space that can be advantageous when finding an update direction. Specifically, to account for all parameter values, we propose to place a probability distribution over the task-specific parameters, which in turn induces a probability distribution over the gradients. As a result, we obtain uncertainty estimates for the gradients that reflect the sensitivity in each of their dimensions. High-uncertainty dimensions are more lenient for changes while dimensions with a lower uncertainty are more strict (see illustration in \Figref{fig:update_dir}).



To obtain a probability distribution over the task-specific parameters we take a Bayesian approach. According to the Bayesian view, a posterior distribution over parameters of interest can be derived through Bayes rule. In MTL, it is common to use a shared feature extractor network with linear task-specific layers \cite{ruder2017overview}.
Hence, if we assume a Bayesian model over the last task-specific layer weights during the back-propagation process, we obtain the posterior distributions over them. The posterior is then used to compute a Gaussian distribution over the gradients by means of moment matching.
Then, to derive an update direction for the shared network, we design a novel aggregation scheme that considers the full distributions of the gradients. We name our method \textit{BayesAgg-MTL}. An important implication of our approach is that \MN assigns weights to the gradients at a higher resolution compared to existing methods, allocating a specific weight for each dimension and datum in the batch. We demonstrate our method effectiveness over baseline methods on the MTL benchmarks QM9 \cite{ramakrishnan2014quantum}, CIFAR-100 \cite{krizhevsky2009learning}, ChestX-ray14 \cite{wang2017chestx}, and UTKFace \cite{zhang2017age}.


In summary, this paper makes the following novel contributions:
(1) The first Bayesian formulation of gradient aggregation for Multi-Task Learning. (2) A novel posterior approximation based on a second-order Taylor expansion. (3) A new MTL optimization algorithm based on our posterior estimation. (4) New state-of-the-art results on several MTL benchmarks compared to leading methods. Our code is publicly available at \textcolor{VioletRed}{\url{https://github.com/ssi-research/BayesAgg_MTL}}.

\section{Background}
\label{sec:background}
\textbf{Notations.} Scalars, vectors, and matrices are denoted with lower-case letters (e.g., $x$), bold lower-case letters (e.g., $\rvx$), and bold upper-case letters (e.g., $\rmX$) respectively. All vectors are treated as column vectors. Training samples are tuples consisting of shared features across all tasks and  labels of $K$ tasks, namely $(\rvx, \{\rvy^k\}_{k=1}^K) \sim \gD$, where $\gD$ denotes the training set. We denote the dimensionality of the input and the output of task $k$ by $d_\rvx$ and $o_k$ accordingly. 

In this study, we focus on common NN architectures for MTL having a shared feature extractor and linear task-specific heads \cite{kendall2018multi, sener2018multi}. 
The model parameters are denoted by $\{\rvtheta, \{\rvw^k\}_{k=1}^K\}$, where $\rvtheta \in \sR^{d_{\rvtheta}}$ is the vector of shared parameters and $\{\rvw^k\}_{k=1}^K$ are task-specific parameter vectors, each lies in $\sR^{d_k}$. The last shared feature representation is denoted by the vector $\rvh(\rvx; \rvtheta) \in \sR^{d_\rvh}$. 
Hence, the output of the network for task $k$  can be described as $\rvf^k(\rvh(\rvx;\rvtheta);\rvw^k)$. 
The loss of task $k \in [1, ..., K]$ is denoted by $\ell^k(\rvx, \rvy;\{\rvtheta, \rvw^k\})$.
The gradient of loss $\ell^k$ w.r.t $\rvh(\rvx;\rvtheta)$ is $\rvg^k \coloneqq \frac{\partial \ell^k}{\partial \rvh(\rvx;\rvtheta)}(\rvx, \rvy;\{\rvtheta,\rvw^k\}) \in\sR^{d_h}$. For clarity of exposition, function dependence on input variables will be omitted from now on.

\begin{figure}[!t]
    \centering
    \includegraphics[width=0.9\linewidth]{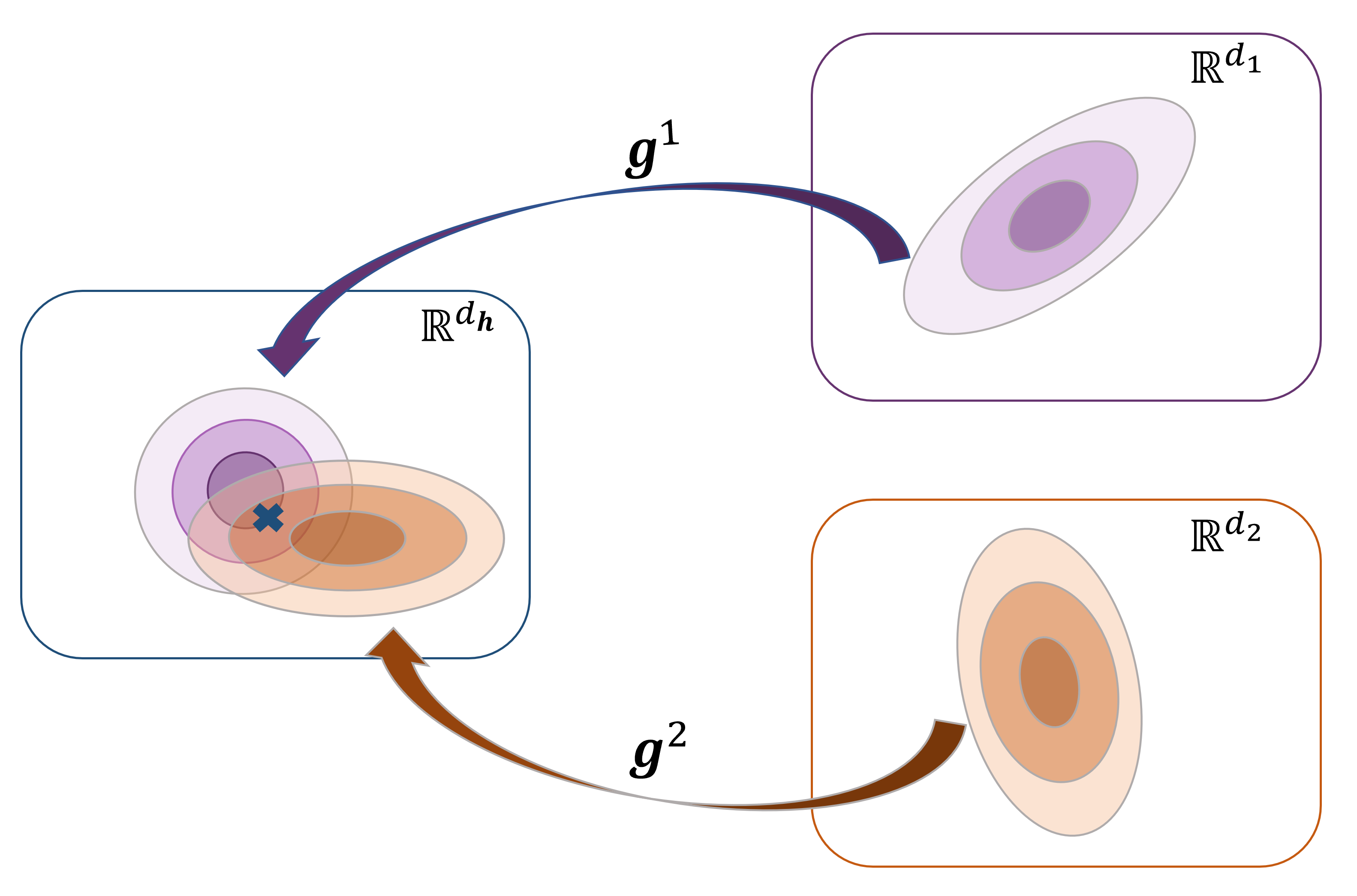}
    \vspace{-0.2cm}
\caption{\MN assumes a probability distribution over the last layer parameters of each task. It first maps these distributions to the space of the last shared representation. Then an update direction is found for the shared representation based on the mean and variance of all distributions (denoted by \textcolor{BlueViolet}{\textbf{X}}).}
\label{fig:dist_mapping}
\end{figure}

\subsection{Multi-Task Learning}
A prevailing approach to optimize MTL models goes as follows. First, the gradient of each task loss is computed. Second, an aggregation rule is imposed to combine the gradients according to some algorithm. And lastly, perform an update step using the outcome of the aggregation step. Commonly the aggregation rule operates on the gradients of the loss w.r.t parameters, or only the shared parameters \citep[e.g.,][]{yu2020gradient, navon2022multi,shamsian2023auxiliary}). Alternatively, to avoid a costly full back-propagation process for each task, some methods suggest applying it on the last shared representation \citep[e.g.,][]{sener2018multi, liu2020towards, senushkin2023independent}. Here, to make our method fast and scalable, we take the latter approach and note that it could be extended to full gradient aggregation.

\subsection{Bayesian Inference}
We wish to incorporate uncertainty estimates for the gradients into the aggregation procedure. Doing so will allow us to find an update direction that takes into account the importance of each gradient dimension for each task. A natural choice to model uncertainty is using Bayesian inference. Since we would like to get uncertainty estimates w.r.t the last shared hidden layer, we treat only the last task-specific layer as a Bayesian layer. This ``Bayesian last layer" approach is a common way to scale Bayesian inference to deep neural networks  \cite{snoek2015scalable, calandra2016manifold, wilson2016deep, achituve2021personalized}. We will now present some of the main concepts of Bayesian modeling that will be used as part of our method.

For simplicity, assume a single output variable. We also dropped the task notation for clarity. According to the Bayesian paradigm, instead of treating the parameters $\rvw$ as deterministic values that need to be optimized, they are treated as random variables, i.e. there is a distribution over the parameters. The posterior distribution for $\rvw$, after observing the data, is given using Bayes rule as 
\begin{equation} 
    \label{eq:post}
    \begin{aligned}
    log~p(\rvw | \gD) &\propto log~p(\rvy| \rmX, \rvw) + log~p(\rvw).
    \end{aligned}
\end{equation}
Predictions in Bayesian inference are given by taking the expected prediction with respect to the posterior distribution. 
In general, the Bayesian inference procedure for $\rvw$ is intractable. However, for some specific scenarios, there exists an analytic solution. For example, in linear regression, if we assume a Gaussian likelihood with a fixed independent scalar noise between the observations $\tau$, $p(\rvy| \{\rvx_i\}_{i=1}^{|\gD|}, \rvw) = \prod_{i=1}^{|\gD|} \normal(y_i|\rvx_i^T \rvw, \tau^2)$, and a Gaussian prior $p(\rvw) = \normal(\rvw|\rvm_p,\rmS_p)$ then, 
\begin{equation} 
    \label{eq:reg_post}
    \begin{aligned}
    p(\rvw | \gD) &= \normal(\rvw | \rvm,\rmS)\\
    \rvm &= \rmS ((\rmS_p)^{-1}\rvm_p + \tau^{-2} \rmX\rvy)\\
    \rmS &= ((\rmS_p)^{-1} + \tau^{-2}\rmX\rmX^T)^{-1}.
    \end{aligned}
\end{equation}
Here $\rmX \in \sR^{d_\rvx \times |\gD|}$ is the matrix that results from stacking the vectors $\{\rvx_i\}_{i=1}^{|\gD|}$. Similarly, we denote by $\rmH \in \sR^{d_\rvh \times |\gD|}$ the matrix that results from stacking the vectors of hidden representation. In the specific case of deep NNs with Bayesian last layer we get the same inference result only with $\rmH$ replacing $\rmX$. Going beyond a single output variable entails defining a covariance matrix for the noise model. However, in this study we assume independence between the output variables in these cases. 

Unlike regression, in classification the likelihood is not a Gaussian, and the posterior can only be approximated. The common choice is to use variational inference \cite{wilson2016stochastic, achituve2021gp, achituve2023guided}, although there are other alternatives as well \cite{kristiadi2020being}.


\section{Method}
\label{sec:method}
We start with an outline of the problem and our approach. Consider a deep network for multi-task learning that has a shared feature extractor part and task-specific linear layers. We propose to use Bayesian inference on the last layer as a means to train \textit{deterministic} MTL models. For each task $k$, we define a Bayesian probabilistic model representing the uncertainty over the linear weights of the last, task-specific layer $\rvw^k$. The distribution over weights induces a distribution over gradients of the loss with respect to the last \textit{shared} hidden layer. 
Given these per-task distributions on a joint space, we propose an aggregation rule for combining the gradients of the tasks to a shared update direction that takes into account the uncertainty in the gradients (see illustration in \Figref{fig:dist_mapping}). Then, the back-propagation process can proceed as usual. 


Since regression and classification setups yield different inference procedures according to our approach, albeit having the same general framework, we discuss the two setups separately, starting with regression.

\subsection{\MN for Regression Tasks}
\label{sec:method_reg}

Consider a standard square loss for task $k$, $\ell^k = (y^k - \hat{y}^k)^2$,  between the label $y^k$ and the network output $\hat{y}^k$. Given a random batch of example $\gB \sim \gD$,  the gradient of the loss with respect to the hidden layer $\rvh$ for the $i^{th}$ example is, 
\begin{equation}
\label{eq:grad}
    \rvg^k_i = \frac{\partial l^k_i}{\partial \hat{y}^k_i}\frac{\partial \hat{y}^k_i}{\partial \rvh_i} = 2 \rvw^k (\rvh^T_i \rvw^k - y^k_i).
\end{equation}
Our main observation is that $\rvg^k_i$ is a function of $\rvw^k$. Hence, if we view $\rvw^k$ in the back-propagation process as a random variable, then $\rvg^k_i$ will be a random variable as well. This view will allow us to capture the uncertainty in the task gradient. Since the dimension of the hidden layer is usually small compared to the dimension of all shared parameters, operations in this space, such as matrix inverse, should not be costly.
 
If we fix all the shared parameters, then the posterior over $\rvw^k$ has a Gaussian distribution with known parameters via \Eqref{eq:reg_post}. As  $\rvg^k_i$ is quadratic in $\rvw^k$, it has a generalized chi-squared distribution \cite{davies1973numerical}. However, since this distribution does not admit a closed-form density function, and since the gradient aggregation needs to be efficient as we run it at each iteration, we approximate $\rvg^k_i$ as a Gaussian distribution. The optimal choice for the parameters of this Gaussian is given by matching its first two moments to those of the true density, as these parameters minimize the Kullback–Leibler divergence between the two distributions \cite{minka2001expectation}. Luckily, in the regression case, we can derive the first two moments from the posterior over $\rvw^k$,
\begin{equation}
\label{eq:explicit_mom}
\begin{aligned} 
    &\E[\rvg^k_i] = 2 [\rmS^k\rvh_i + \rvm^k(\rvh_i^T \rvm^k - y^k_i)],\\
    &\E[\rvg^k_i(\rvg^k_i)^T] =  
    4[(y^k_i)^2(\rmS^k + \rmM^k) - 2y^k_i(\rvm^k\rvh_i^T(\rmS^k + \rmM^k)\\
    &+ (\rmS^k + \rmM^k)\rvh_i(\rvm^k)^T
     + \rvh_i^T\rvm^k(\rmS^k - \rmM^k))\\
    &+ (\rmS^k + \rmM^k)(\rmA_i + \rmA_i^T)(\rmS^k + \rmM^k)\\
    &+ Tr(\rmA_i\rmS^k)(\rmS^k + \rmM^k)+ (\rvm^k)^T\rmA_i\rvm^k(\rmS^k - \rmM^k)],
\end{aligned}
\end{equation}
where $\rmA_i = \rvh_i\rvh_i^T$, $\rmM^k = \rvm^k(\rvm^k)^T$, we assumed $\tau = 1$, and $Tr(\cdot)$ is the matrix trace. We emphasize that the following approximation is for the gradient of a single data point and a single task, not for the gradient of the task with respect to the entire batch.  The full derivation is presented in \Appref{sec:full_derivations_reg}.

Several points deserve attention here. First, note the similarity between the solution of the first moment and the gradient obtained via the standard back-propagation. The two differences are that the last layer parameters, $\rvw^k$, are replaced with the posterior mean, $\rvm^k$, and an uncertainty term was added. In the extreme case of $\rmS^k \rightarrow 0$ and $\rvm^k \rightarrow \rvw^k$, the mean coincides with that of the standard back-propagation. Second, in the case of a multi-output task, following our independence assumption between output variables, we can obtain the moments for each output dimension separately using the same procedure, so de facto we treat each output as a different task.  Finally, during training, the shared parameters are constantly being updated. Hence, to compute the posterior distribution for $\rvw^k$ we need to iterate over the entire dataset at each update step. In practice, this can make our method computationally expensive. Therefore, we use the current batch data only to approximate the posterior over $\rvw^k$, and introduce information about the full dataset through the prior as described next.

\textbf{Prior selection.} 
A common choice in Bayesian deep learning is to choose uninformative priors, such as a standard Gaussian, to let the data be the main influence on the posterior \cite{wilson2020bayesian, fortuin2021bayesian}. However, in our case, we found this prior to be too weak. Since the posterior depends only on a single batch we opted to introduce information about the whole dataset through the prior. 
A natural choice is to use the posterior distribution of the previous batch as our prior \citep[Chapter~3]{SimoBayes}. However, this method did not work well in our experiments and we developed an alternative. During each epoch, we collect the feature representations and labels of all examples in the dataset. At the end of the epoch, we compute the posterior based on the full data (with an isotropic Gaussian prior) and use this posterior as the prior at each step in the subsequent epoch. Updating the full data prior more frequently is likely to have a beneficial effect on our overall model; however it will also probably make the training time longer. Hence, doing the update once an epoch strikes a good balance between performance and training time.

\textbf{Aggregation step.} Having an approximation for the gradient distribution of each task we need to combine them to find an update direction for the shared parameters. Denote the mean of the gradient of the loss for task $k$ w.r.t the hidden layer for the $i^{th}$ example by $\rvmu^k_i \coloneqq \E[\rvg^k_i]$, and similarly the covariance matrix $\rmSigma^k_i \coloneqq (\rmLambda^k_i)^{-1} \coloneqq \E[\rvg^k_i(\rvg^k_i)^T] - \E[\rvg^k_i]\E[\rvg^k_i]^T$. We strive to find an update direction for the last shared layer, $\rvg_i$, that lies in a high-density region for all tasks. Hence, we pick $\rvg_i$ that maximizes the following likelihood:
\begin{equation}
\label{eq:nll_update}
\begin{aligned} 
    \argmax_{\rvg_i} &\prod_{k=1}^K \normal(\rvg_i | \rvmu^k_i, \rmSigma^k_i) = \\
    \argmin_{\rvg_i} &- \sum_{k=1}^K log~\normal(\rvg_i | \rvmu^k_i, \rmSigma^k_i).
\end{aligned}
\end{equation}
Thankfully, the above optimization problem can be solved in closed-form, yielding the following solution:
\begin{equation}
\label{eq:nll_solution_full}
\begin{aligned} 
    \rvg_i = \left(\sum_{k=1}^K \rmLambda^k_i\right)^{-1}\left(\sum_{k=1}^K \rmLambda^k_i\rvmu^k_i\right).
\end{aligned}
\end{equation}
However, we found that modeling the full covariance matrix can be numerically unstable and sensitive to noise in the gradient. Instead, we assume independence between the dimensions of $\rvg_i^k$ for all tasks which results in diagonal covariance matrices having variance $(\rvsigma^k_i)^2 \coloneqq 1 / \rvlambda^k_i$. The update direction now becomes:
\begin{equation}
\label{eq:nll_solution_diag}
\begin{aligned} 
    \rvg_i = \sum_{k=1}^K\frac{1/(\rvsigma^k_i)^2 }{\sum_{k=1}^K 1/(\rvsigma^k_i)^2} \rvmu^k_i = \sum_{k=1}^K \overbrace{\frac{\rvlambda^k_i }{\sum_{k=1}^K \rvlambda^k_i}}^{\rvalpha_i^k}\rvmu^k_i,
\end{aligned}
\end{equation}
where the division and multiplication are done element-wise. In \Eqref{eq:nll_solution_diag} we intentionally denote by $\rvalpha_i^k$ the vector of uncertainty-based weights that our method assigns to the mean gradient to highlight that the weights are unique per task, dimension, and datum. The final modification for the method involves down-scaling the impact of the precision by a hyper-parameter $s \in (0,1]$, namely, we take $(\rvlambda^k_i)^s$. Empirically, the scaling parameter helped to achieve better performance, perhaps due to misspecifications in the model (such as the diagonal Gaussian assumption over $\rvg^k_i$). 

\begin{algorithm}[!t] 
    \caption{ \MN}
    {\bf Input}: $\gB$ - a random batch of examples; $p(\rvw^k|\gD) ~~\forall k \in [1,...,K]$ - posterior distributions over the task-specific parameters; $s$ - scaling hyper-parameter\\
    {\bf For} $i = 1, ..., |\gB|$:\\
    \hspace*{3mm} {\bf For} $k = 1, ..., K$:\\
    \hspace*{6mm} $\bullet$ Compute $\E[\rvg^k_i]$ and $\E[\rvg^k_i(\rvg^k_i)^T]$ as in \Eqref{eq:explicit_mom} for\\
    \hspace*{6mm} regression or \Eqref{eq:moments_mc} for classification.\\
    \hspace*{6mm} $\bullet$ Set (operations are done element-wise),\\
    \hspace*{8.5mm} $\rvmu^k_i \coloneqq \E[\rvg^k_i]$,\\
    \hspace*{8.5mm} $\rvlambda^k_i \coloneqq (\E[(\rvg^k_i)^2] - \E[\rvg^k_i]\E[\rvg^k_i]))^{-1}$.\\
    \hspace*{3mm}{\bf End for}\\
    \hspace*{3mm} Compute $\rvg_i =\sum_{k=1}^K \frac{(\rvlambda^k_i)^s}{\sum_{k=1}^K (\rvlambda^k_i)^s}\rvmu^k_i$.\\   
    {\bf End for}\\
    Compute gradient via matrix multiplication w.r.t the shared parameters: $\frac{1}{|\gB|} \sum_{i=1}^{|\gB|} \rvg_i \frac{\partial \rvh_i}{\partial \rvtheta}$. 
    \label{algo:BayesAggMTL}
\end{algorithm}

\begin{figure}[!t]
    \centering
    \includegraphics[width=0.9\linewidth]{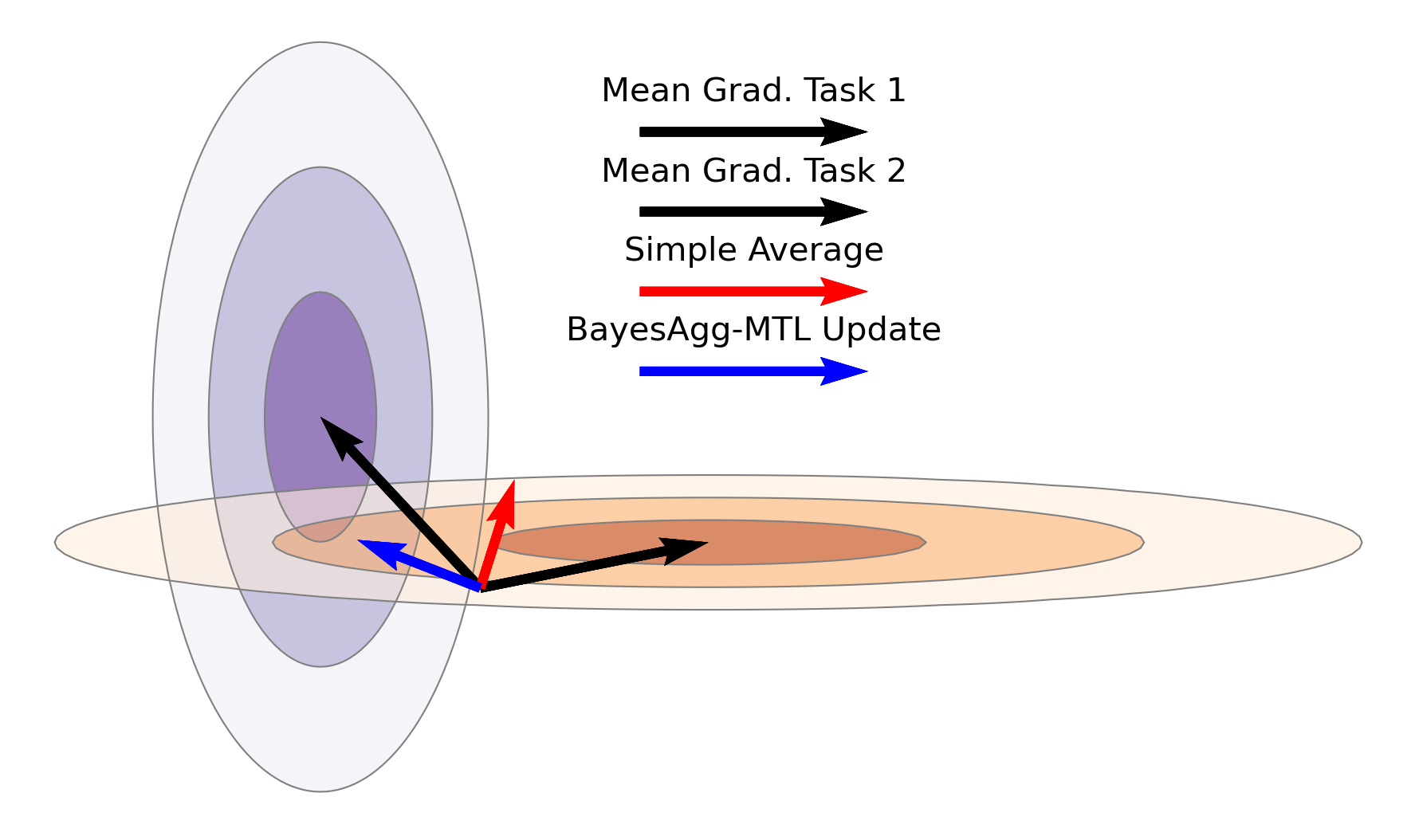}
    \vspace{-0.2cm}
\caption{\MN update for a two-dimensional feature representation. Black arrows indicate the mean update direction of each task; \textcolor{red}{Red} arrow is the update direction of a simple average; \textcolor{blue}{Blue} arrow is the proposed update direction. Darker colors in the contours represent regions with higher density.}
\label{fig:update_dir}
\end{figure}

With the aggregated gradient for each example, the back-propagation procedure proceeds as usual by averaging over all examples in the batch and then back-propagating this over to the shared parameters. To gain a better intuition about the update rule of \MN, consider the illustration in \Figref{fig:update_dir}. In the figure, we plot the mean update direction of two tasks along with the uncertainty in them. The first task is more sensitive to shifts in the vertical dimension and less so to shifts in the second (horizontal) dimension, while for the second task, it is the opposite. By taking the variance information into account, \MN can find an update direction that works well for both, compared to a simple average of the gradient means. We summarize our method in Algorithm \ref{algo:BayesAggMTL}.

\textbf{Making predictions.} Since we have a closed-form solution for the posterior of the task-specific parameters, \MN does not learn this layer during training. Therefore, when making predictions we use the posterior mean, $\rvm^k$, computed on the full training set. We do so, instead of using a full Bayesian inference, for a fair comparison with alternative MTL approaches and to have an identical run-time and memory requirements when making predictions. 

\textbf{Connection to Nash-MTL.}
In \cite{navon2022multi} the authors proposed a cooperative bargaining game approach to the gradient aggregation step with the directional derivative as the utility of each player (task). They then proposed using the Nash bargaining solution, the direction that maximizes the product of all the utilities. One can consider \Eqref{eq:nll_update} as the Nash bargaining solution with the utility of each task being its likelihood. However, unlike \cite{navon2022multi} we get an analytical formula for the bargaining solution since the Gaussian exponent and the logarithm cancel out.  

\subsection{\MN for Classification Tasks}
\label{sec:method_cls}
We now turn to present our approach for classification tasks. When dealing with classification there are two sources of intractability that we need to overcome. The first is the posterior of $\rvw^k$, and the second is estimating the moments of $\rvg^k_i$. We describe our solution to both challenges next.
 
\textbf{Posterior approximation.} In classification tasks the likelihood is not a Gaussian and in general, we cannot compute the posterior in closed-form. One common option is to approximate it using a Gaussian distribution and learn its parameters using a variational inference (VI) scheme \cite{saul1996mean, neal1998view,bishop2006pattern}. However, in our early experimentations, we didn't find it to work well without using a computationally expensive VI optimization at each update step. Alternatively to VI, the Laplace approximation \cite{mackay1992bayesian} approximates the posterior as a Gaussian using a second-order Taylor expansion. Since the expansion is done at the optimal parameter values that are learned point-wise, the Jacobean term in the expansion vanishes. Here, we follow a similar path; however, we cannot assume that the Jacobean is zero as we are not near a stationary point during most of the training. Nevertheless, we can still find a Gaussian approximation. A similar derivation was proposed in \cite{immer2021scalable}, yet they ignored the first order term eventually.
Denote by $\hat{\rvw}^k$ the \textit{learned} point estimate for the task parameters, and  $\Delta \rvw^k \coloneqq \rvw^k - \hat{\rvw}^k$. Then, at each step of the training by using Bayes rule we can obtain a posterior approximation for $\rvw^k$ using the following:
\begin{equation}
\label{eq:taylor_exp}
\begin{aligned} 
    &log~p(\rvw^k | \gB) \approx log~p(\hat{\rvw}^k | \gB) + \\
    & \left(- \frac{\partial log~p(\rvy^k | \rmX, \rvw^k)}{\partial \rvw^k} -  \frac{\partial log~p(\rvw^k)}{\partial \rvw^k}\right)^T  \Delta \rvw^k + \\
    &\frac{1}{2} (\Delta \rvw^k)^T \left(- \frac{\partial^2 log~p(\rvy^k | \rmX, \rvw^k)}{\partial (\rvw^k)^2}
    - \frac{\partial^2 log~p(\rvw^k)}{\partial (\rvw^k)^2} \right) \Delta \rvw^k. \\
\end{aligned}
\end{equation}
The above takes the following form $c^k + (\rva^k)^T (\rvw^k - \hat{\rvw}^k) + \frac{1}{2} (\rvw^k - \hat{\rvw}^k)^T \rmB^k (\rvw^k - \hat{\rvw}^k)$, where $\rva^k \in \sR^{d_k}, \rmB^k \in \sR^{d_k \times d_k}, c^k \in \sR$ are known constants. We stress here again, that since we apply Bayesian inference to the last layer parameters only, computing and inverting $\rmB^k$, typically does not incur a large computational overhead. 

After rearranging and completing the square we obtain a quadratic form corresponding to the following Gaussian distribution (see full derivation in \Appref{sec:full_derivations_cls}):
\begin{equation}
\label{eq:post_cls}
\begin{aligned} 
    p(\rvw^k | \gB) &\approx \normal(\rvw^k | \hat{\rvw}^k - (\rmB^k)^{-1}\rva^k, (\rmB^k)^{-1}).
\end{aligned}
\end{equation}
Examining the above posterior reveals several insights. First, the posterior mean corresponds to the Newton method update step. Second, the covariance of this posterior is the same as that of the Laplace approximation. Third, at a stationary point the Laplace approximation is recovered if the gradient of the loss w.r.t the parameters approaches zero.

One limitation of the approximation in \Eqref{eq:post_cls} is that the Hessian will not be positive-definite in most cases. Therefore, we replace it with the generalized Gauss-Newton (GGN) matrix \cite{schraudolph2002fast, martens2011learning, daxberger2021laplace}:
\begin{equation}
\begin{aligned} 
    \tilde{\rmB}^k = \sum_{i=1}^{|\gB|} (\rmJ^k_i)^T \rmH^k_i \rmJ^k_i + (\rmS_p^k)^{-1}.
\end{aligned}
\end{equation}
Where, $\rmJ^k_i = \partial \rvf^k(\rvx_i;\rvw^k)/\partial \rvw^k \in \sR^{o_k \times d_k}$ is the Jacobean of the model output for task $k$ w.r.t the last layer parameters of that task, $\rmH^k_i = - \partial^2 log~p(\rvy^k_i | \rvx_i, \rvw^k) / \partial (\rvf^k(\rvx_i;\rvw^k))^2 \in \sR^{o_k \times o_k}$ is the Hessian of the negative log-likelihood w.r.t the model outputs of task $k$, and $\rmS_p^k$ is the covariance of the Gaussian prior for $\rvw^k$. As in the regression case we use here an informative prior based on the posterior from the full dataset at each training step.

\textbf{Moments estimation.} Unlike the regression case, in classification $\rvg^k_i$ will depend on $\rvw^k$ through some non-linear function. Hence, obtaining the moments as in \Eqref{eq:explicit_mom} in closed-form is more challenging. However, since we are estimating the parameters of the last layer only, which in many cases are relatively low-dimensional, we can efficiently approximate these moments with Monte-Carlo sampling:
\begin{equation}
\label{eq:moments_mc}
\begin{aligned} 
    \E[\rvg_i^k] &\approx  \frac{1}{J} \sum_{j=1}^J \rvg_i^k(\rvw^k_j),\\
    \E[\rvg_i^k(\rvg_i^k)^T] &\approx \frac{1}{J} \sum_{j=1}^J \rvg_i^k(\rvw^k_j) \rvg_i^k(\rvw^k_j)^T.
\end{aligned}
\end{equation}
Here, $\rvw_j^k$ are samples from $p(\rvw^k | \gB)$, and the total number of samples are $J$. Effectively this means that we need to back-propagate gradients w.r.t the shared hidden layer $J$ times; however, since the task-specific layers are linear it can be done cheaply and in parallel. Having the moment estimation we proceed with the aggregation rule as described in \Secref{sec:method_reg}.


\textbf{Making predictions.} Unlike the regression case, here we learn the parameters of the last layer as part of the posterior approximation. Therefore, making predictions is done as usual with a forward-pass through the network.

\section{Related Work}
\label{sec:related_work}
Multi-task learning is an active research area that attempts to learn jointly multiple tasks, commonly using a shared representation \cite{ruder2017overview, navon2022multi, liu2023famo, elich2023challenging, shi2023deep, Yun_2023_ICCV}.  
Learning a shared representation for multiple tasks imposes some challenges. One challenge is trying to learn an architecture that can express both task-shared and task-specific features. Another challenge is to find the optimal balancing of the tasks and enable learning the different tasks with equal importance.
One line of research in MTL suggests methods to introduce novel MTL-friendly architectures, such as task-specific modules \cite{CrossStitch_2016}, attention-based networks \cite{liu2019end}, and an ensemble of single-task models \cite{dimitriadis2023pareto}. Yet, a more common line of research focuses on the MTL optimization process, trying to explain the difficulties in the process by e.g. conflicting gradients \cite{wang2020gradient} or plateaus in the loss landscape \cite{schaul2019ray}. Our method focuses on the latter, MTL optimization process improvement.


Different strategies were proposed to address the MTL optimization challenge to successfully balance the training of the different tasks and resolve their conflicts. The methods can broadly be categorized into two groups, loss-based and gradient-based \cite{dai2023improvable}. Loss-based approaches attempt to allocate weights for the tasks based on some criteria related to the loss, such as the difficulty of the task \cite{DTP_Guo_2018_ECCV}, 
random weights \cite{lin2022reasonable}, geometric mean of the task losses \cite{GLS2019_CVPR, Yun_2023_ICCV}, 
and task uncertainty \cite{kendall2018multi}. Regarding the last one, to weigh the tasks it uses the uncertainty in the observations \textit{only}. This is very different from our approach that weighs each dimension of the task \textit{gradients} based on full Bayesian information. 

\begin{figure}[!t]
    \centering
    \includegraphics[width=0.9\linewidth]{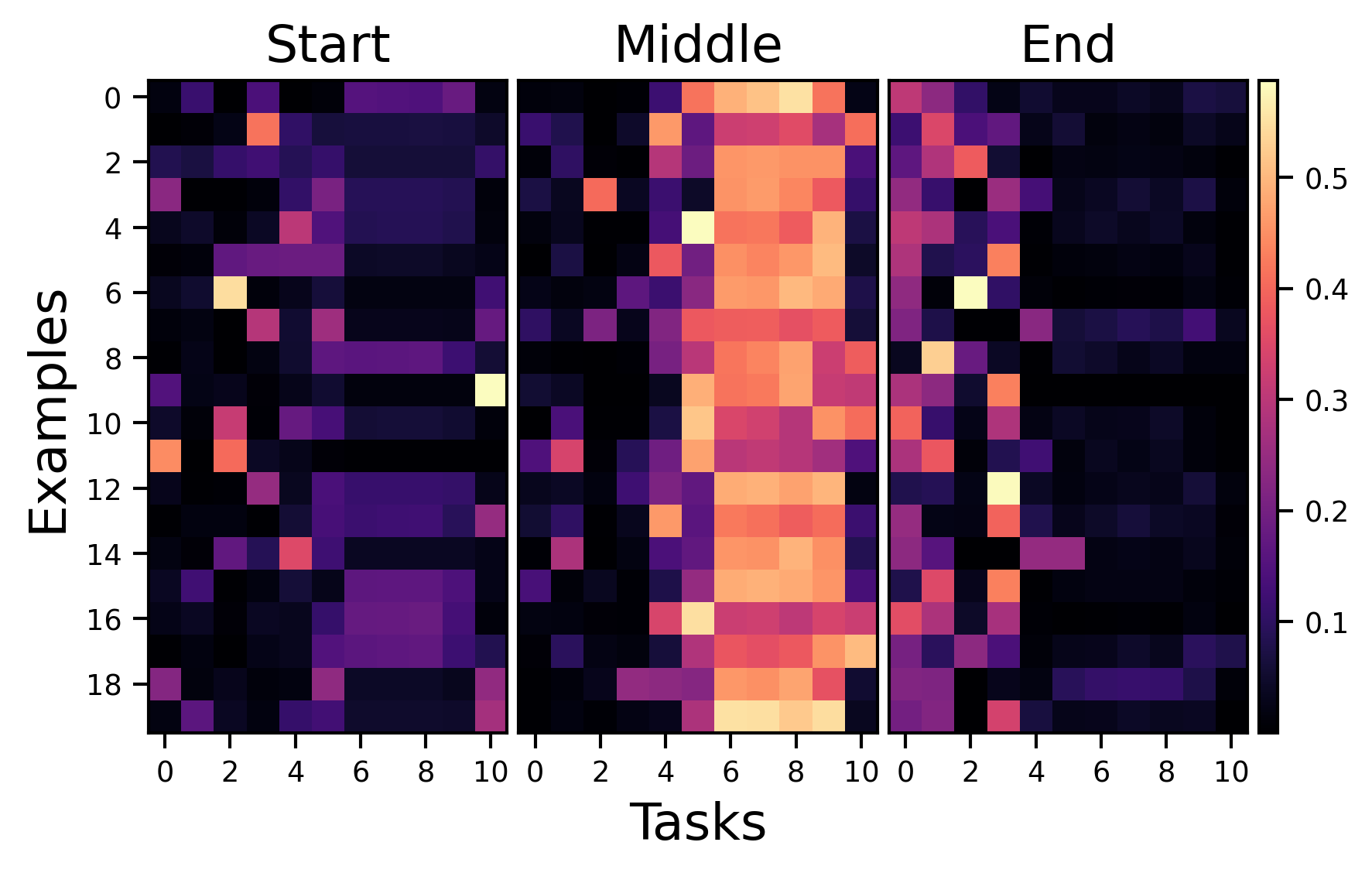}
    \vspace{-0.2cm}
\caption{Mean weight over dimensions per-example for $20$ random examples on the QM9 dataset at different training stages.}
\label{fig:mean_weight}
\end{figure}


Gradient-based methods attempt to balance the tasks by using the gradients information directly \cite{GradNorm2018, NEURIPS2020_GradDrop, javaloy2022rotograd, liu2020towards, navon2022multi, Fernando2023MitigatingGB, senushkin2023independent}. For example, GradNorm \cite{GradNorm2018} dynamically tunes the gradient magnitudes to prevent imbalances between the tasks during training. PCGrad \cite{yu2020gradient} identifies gradient conflicts as the main optimization issue in MTL, and attempts to reduce the conflicts by projecting each gradient to the other tasks' normal plane.
Nash-MTL \cite{navon2022multi} suggests treating MTL as a bargaining game to find Pareto optimal solutions.
Several studies suggested adaptations for the multiple-gradient descent algorithm (MGDA) \cite{desideri2012multiple, sener2018multi}, such as CAGrad, \cite{liu2021conflict}, and MoCo \cite{Fernando2023MitigatingGB}.
As opposed to previous methods, our approach considers both the mean and the variance of the gradients to derive an update direction.

Lastly, some studies recently suggested performing model merging based on the uncertainty of the parameters \cite{matena2022merging, daheim2023model}. The goal there is usually to combine models for various tasks, such as model ensembling, federated learning, and robust fine-tuning. Unlike these methods, we assume a Bayesian model on the last layer only and propagate the uncertainty to the gradients for gradient aggregation.

\begin{table}[!t]
\centering
\footnotesize
\caption{\textit{QM9}. Test performance averaged over 3 random seeds.}
    \vskip 0.11in
\begin{tabular}{@{}llcc@{}}
\toprule
 & & $\mathbf{\Delta_m\%}$ ($\downarrow$) \\ 
 \midrule
\multicolumn{2}{c}{LS} & $177.6 \pm 3.4$ \\
\multicolumn{2}{c}{SI} & $~~77.8 \pm 9.2$ \\
\multicolumn{2}{c}{RLW} & $203.8 \pm 3.4$ \\
\multicolumn{2}{c}{DWA} & $175.3 \pm 6.3$ \\
\multicolumn{2}{c}{UW}  & $~108.0 \pm 22.5$\\
\multicolumn{2}{c}{MGDA} & $120.5 \pm 2.0$\\
\multicolumn{2}{c}{PCGrad} & $~~125.7 \pm 10.3$ \\
\multicolumn{2}{c}{CAGrad} & $112.8 \pm 4.0$ \\ 
\multicolumn{2}{c}{IMTL-G} & $~~77.2 \pm 9.3$ \\ 
\multicolumn{2}{c}{Nash-MTL} & $~~62.0 \pm 1.4$ \\
\multicolumn{2}{c}{IGBv2} & $~~67.7 \pm 8.1$ \\
\multicolumn{2}{c}{Aligned-MTL-UB} & $~~71.0 \pm 9.6$ \\  %
\midrule
\multicolumn{2}{c}{\MN (Ours)} & $\mathbf{~~53.2 \pm 7.1}$\\
\bottomrule
\end{tabular}
\label{tab:qm9}
\end{table}

\section{Experiments}
\label{sec:experiments}
We evaluated \MN on several MTL benchmarks differing in the number of tasks and their types. Unless specified otherwise, we report the average and standard deviation (std) of relevant metrics over $3$ random seeds. In all datasets, we pre-allocated a validation set from the training set for hyper-parameter tuning and early stopping for all methods. Throughout our experiments, we used the ADAM optimizer \cite{KingmaB14} which was found to be effective for MTL due to partial loss-scale invariance \cite{elich2023challenging}. Full experimental details are given in \Appref{sec:full_exp_details}. 

\textbf{Compared methods.} We compare \MN with the following baseline methods: \textbf{(1) Single Task Learning (STL)}, which learns each task independently under the same experimental setup as that of the MTL methods; \textbf{(2) Linear Scalarization (LS)}, which assigns a uniform weight to all tasks, namely $\sum_{k=1}^K \ell^k$; \textbf{(3) Scale-Invariant (SI) }\cite{navon2022multi}, which assigns a uniform weight to the log of all tasks, namely $\sum_{k=1}^K log~\ell^k$;
\textbf{(4) Random Loss Weighting (RLW) }\cite{lin2022reasonable}, which allocates random weights to the losses at each iteration; 
\textbf{(5) Dynamic Weight Average (DWA)} \cite{liu2019end}, which allocates a weight based on the rate of change of the loss for each task; \textbf{(6) Uncertainty weighting (UW)} \cite{kendall2018multi}, which minimize a scalar term corresponding to the \textit{aleatoric} uncertainty for each task; \textbf{(7) Multiple-Gradient Descent Algorithm (MGDA)} \cite{desideri2012multiple, sener2018multi}, which finds a minimum norm solution for a convex combination of the losses; \textbf{(8) Projecting Conflicting Gradients (PCGrad)} \cite{yu2020gradient}, which projects the gradient of each task onto the normal plane of tasks they are in conflict with; \textbf{(9) Conflict-Averse Grad (CAGrad) }\cite{liu2021conflict}, which searches an update direction centered at the LS solution while minimizing conflicts in gradients; \textbf{ (10) Impartial MTL-Grad (IMTL-G)} \cite{liu2020towards}, which finds an update vector such that the projection of it on each of the gradients of the tasks is equal; \textbf{(11) Nash-MTL \cite{navon2022multi}} that derives task weights based on the Nash bargaining solution; \textbf{(12) Improvable Gap
Balancing (IGBv2)} \cite{dai2023improvable}, which suggests a Reinforcement learning procedure to balance the task losses; \textbf{(13) Aligned-MTL-UB \cite{senushkin2023independent}}, which aligns the principle components of a gradient matrix. 

\textbf{Evaluation metric.} Unless specified otherwise, we report the $\Delta_m\%$ metric introduced in \cite{maninis2019attentive}. This metric measures the average relative difference between a method $m$ compared to the STL baseline according to some criterion of interest $M^k$. Namely, $\Delta_m = \frac{1}{K} \sum_{k=1}^K (-1)^{\delta_k}(M_m^k - M_s^k) / M_s^k$. Where, $M_m^k$ is the criterion value for task $k$ under method $m$, $M_s^k$ is the criterion value for task $k$ under the STL baseline, and $\delta_k \in \{0, 1\}$. If $\delta_k = 0$ then a lower value for $M^k$ is better (e.g., task loss), and if $\delta_k = 1$ then a higher value for $M^k$ is preferred (e.g., task accuracy). Lower $\Delta_m\%$ indicates a better performance.

\textbf{Pre-training stage.} 
To obtain meaningful features for the Bayesian layer, it is a common practice to apply a pre-training step using standard NN training for several epochs \cite{wilson2016deep, wilson2016stochastic}. We follow the same path here and apply an initial pre-training step using linear scalarization. We would like to stress here that in all the experiments, the overall number of training steps for \MN (including the pre-training) is the same as all methods.

\begin{table}[!t]
\footnotesize
\centering
\caption{Test performance averaged over 3 random seeds on binary classification tasks from CIFAR-MTL \& ChestX-ray14 datasets.
}
\vskip 0.11in
\resizebox{\columnwidth}{!}{\begin{tabular}{@{}lcc@{}}
\toprule
    &  \textbf{CIFAR} (Acc.) [$\uparrow$] & \textbf{CX-ray} ($\mathbf{\Delta_m\%}$) [$\downarrow$]\\ 
 \midrule
\multicolumn{1}{c}{LS}  & $56.96 \pm .06$ & $-14.62 \pm 0.2$ \\
\multicolumn{1}{c}{SI}  & $55.75 \pm 0.3$ & $-10.94 \pm 0.4$ \\
\multicolumn{1}{c}{RLW} & $59.30 \pm .08$ & $-11.69 \pm 0.1$ \\
\multicolumn{1}{c}{DWA} & $58.44 \pm 0.5$ & $\mathbf{-14.79 \pm .07}$ \\
\multicolumn{1}{c}{UW}  & $56.63 \pm 0.5$ & $-13.95 \pm 0.2$\\
\multicolumn{1}{c}{MGDA} & $\mathbf{59.74 \pm .07}$ & $-14.44 \pm 0.4$\\
\multicolumn{1}{c}{PCGrad} & $56.32 \pm 0.2$ & $-13.43 \pm 0.5$ \\
\multicolumn{1}{c}{CAGrad} & $56.59 \pm 0.2$ & $-14.49 \pm 0.1$ \\ 
\multicolumn{1}{c}{IMTL-G} & $57.09 \pm 0.3$ & $~-8.23 \pm 1.8$ \\ 
\multicolumn{1}{c}{Nash-MTL} & $56.59 \pm 0.2$ & $-13.23 \pm 0.5$ \\
\multicolumn{1}{c}{IGBv2} & $56.61 \pm 0.2$ & $~-2.82 \pm 0.6$ \\
\multicolumn{1}{c}{Aligned-MTL-UB} & $56.57 \pm 0.7$ & $-14.14 \pm 0.2$ \\  %
\midrule
\multicolumn{1}{c}{\MN (Ours)} & $\mathbf{59.97 \pm 0.4}$ & $\mathbf{-14.96 \pm 0.1}$ \\
\bottomrule
\end{tabular}}
\label{tab:cifar_chest}
\end{table}

\begin{table*}[!t]
\footnotesize
\centering
\caption{\textit{UTKFace}. Test performance averaged over 8 random seeds.}
    \vskip 0.11in
\begin{tabular}{@{}clccccc@{}}
\toprule
 &  & \textbf{Age} \tiny{($\mathbf{\times 10^{1}}$)} \small{($\downarrow$)} &  \textbf{Gender} ($\uparrow$) & \textbf{Ethnicity} ($\uparrow$) & $\mathbf{\Delta_m\%}$ ($\downarrow$) \\
 \cmidrule(lr){3-5}
 \multicolumn{2}{c}{STL} & $1.40 \pm 0.03$ & $92.32 \pm 0.35$ & $82.42 \pm 0.42$ & -- \\
 \midrule
\multicolumn{2}{c}{LS}   & $1.46 \pm 0.02$ & $92.92 \pm 0.24$ & $83.98 \pm 0.43$ & $~~~0.69 \pm 0.59$ \\
\multicolumn{2}{c}{SI}   & $1.42 \pm 0.03$ & $93.05 \pm 0.29$ & $83.40 \pm 0.27$ & $~~~0.11 \pm 0.89$ \\
\multicolumn{2}{c}{RLW}  & $1.44 \pm 0.03$ & $92.89 \pm 0.25$ & $83.70 \pm 0.49$ & $-0.31 \pm 0.76$\\
\multicolumn{2}{c}{DWA}  & $1.44 \pm 0.02$ & $92.90 \pm 0.16$ & $83.55 \pm 0.33$ & $~~~0.35 \pm 0.60$\\
\multicolumn{2}{c}{UW}  & $1.43 \pm 0.00$ & $92.99 \pm 0.24$ & $83.09 \pm 0.39$ & $~~~0.15 \pm 0.24$\\
\multicolumn{2}{c}{MGDA}  & $1.38 \pm 0.02$ & $\mathbf{93.29 \pm 0.31}$ & $83.51 \pm 0.30$ & $-1.39 \pm 0.50$\\
\multicolumn{2}{c}{PCGrad} & $1.47 \pm 0.03$ & $92.92 \pm 0.28$ & $83.28 \pm 0.38$ & $~~~1.13 \pm 0.57$ \\
\multicolumn{2}{c}{CAGrad} & $1.40 \pm 0.02$ & $93.06 \pm 0.26$ & $83.28 \pm 0.46$ & $-0.58 \pm 0.59$ \\ 
\multicolumn{2}{c}{IMTL-G} & $1.41 \pm 0.03$ & $93.10 \pm 0.16$ & $83.78 \pm 0.47$ & $-0.50 \pm 0.89$ \\ 
\multicolumn{2}{c}{Nash-MTL} & $1.42 \pm 0.02$ & $92.89 \pm 0.10$ & $83.19 \pm 0.50$ & $-0.17 \pm 0.71$ \\ 
\multicolumn{2}{c}{IGBv2} & $1.42 \pm 0.02$ & $93.09 \pm 0.22$ & $83.34 \pm 0.33$ & $-0.21 \pm 0.50$ \\
\multicolumn{2}{c}{Aligned-MTL-UB} & $1.45 \pm 0.02$ & $93.00 \pm 0.24$ & $83.36 \pm 0.43$ & $~~~0.66 \pm 0.50$ \\ 
\midrule
\multicolumn{2}{c}{\MN (Ours)} & $\mathbf{1.35 \pm 0.03}$ & $93.01 \pm 0.17$ & $\mathbf{84.25 \pm 0.35}$ & $\mathbf{-2.23 \pm 0.76}$ \\ 
\bottomrule
\end{tabular}
\label{tab:utkface}
\end{table*}

\subsection{\MN for Regression}
\label{sec:qm9_exp}
We first evaluated \MN on an MTL problem with regression tasks only. We used the QM9 dataset which contains $\sim 130,000$ stable small organic molecules represented as graphs having node and edge features \cite{ramakrishnan2014quantum, wu2018moleculenet}. The goal here is to predict $11$ chemical properties, such as geometric and energetic ones, that may vary in scale and difficulty of the tasks. We follow the experimental protocol of \citet{navon2022multi}. Specifically, we allocate approximately $110,000$ examples for training, with separate validation and testing sets with $10,000$ examples each. Additionally, we employ the message-passing neural network architecture \cite{gilmer2017neural} in conjunction with the pooling operator described in \cite{VinyalsBK15}.

The test results for this dataset are presented in \Tableref{tab:qm9}. Baseline method results were taken from \cite{dai2023improvable}, except for Aligned-MTL-UB, which is included here for the first time. The criterion used in $\Delta_m$ here is the mean absolute error (MAE) of the losses. From the table, \MN achieves the best test performance, with a significant improvement compared to most of the baseline methods.

To gain a better intuition into the weights that \MN assigns, we define here again the vector of weights per example and task from \Eqref{eq:nll_solution_diag}, $\rvalpha^k_i \coloneqq {\rvlambda^k_i } / {(\sum_{k=1}^K \rvlambda^k_i})$.  \Figref{fig:mean_weight} depicts for all tasks the average over dimensions of $\rvalpha^k_i$ for $20$ random examples at the start, middle, and end of training. The plot reveals an interesting pattern. Early in training, the average weights are distributed among the tasks without any specific pattern. As training progresses, larger weights are assigned for tasks $4-10$ in the middle of the training, while tasks $0-3$ receive smaller weights. At the end of the training, this pattern changes, and tasks $0-3$ are assigned with larger weights compared to tasks $4-10$.

\subsection{\MN for Binary Classification}
\label{sec:cifar_chest_exp}
Next, we evaluated \MN on the MTL benchmarks CIFAR-MTL \cite{krizhevsky2009learning, rosenbaum2018routing}, and ChestX-ray14 \cite{wang2017chestx}. To the best of our knowledge, we are the first to evaluate MTL methods on the latter dataset. These datasets contain a large number of tasks, $20$ and $14$ respectively, with a high class-imbalance distribution. This poses a significant challenge for current MTL methods.

CIFAR-MTL uses the coarse labels of the CIFAR-100 dataset to create an MTL benchmark having $20$ binary tasks. Classes from this dataset are grouped into super-classes (fish, flowers, trees, etc.), such that each example is given a one-hot encoding vector of labels indicating the super-class it belongs to. We use the official train-test split having $50,000$ examples and $10,000$ examples respectively. We allocate $5,000$ examples from the training set for a validation set. Our experiments on this dataset were conducted using a simple NN having $3$ convolution layers. 

ChestX-ray14 contains $\sim 112,000$ X-ray images of chests from $32,717$ patients. Each image has labels from $14$ binary classes corresponding to the occurrence or absence of thoracic diseases. Multiple diseases can appear together in a patient. In our experiments, we mostly follow the training protocol suggested in \cite{taslimi2022swinchex} that used ResNet-34 for the shared parameters. 
we use the official split of $70\% - 10\% - 20\%$  for training, validation, and test. 

We present the test results for these datasets in \Tableref{tab:cifar_chest}. On the CIFAR-MTL we report the accuracy in class assignment, and on the ChestX-ray14 we report the $\Delta_m$ based on the AUC-ROC values per task. From the table, \MN performs best on both datasets. Interestingly, on the ChestX-ray14 dataset almost all methods, except for ours and DWA, under-perform the naive LS baseline. In \Appref{sec:train_time} we compare the run-time of all methods on this dataset and on the QM9. We show that \MN is substantially faster than other baseline methods that use gradients w.r.t the shared parameters to weigh the tasks.

\subsection{\MN for Mixed Tasks}
\label{sec:utkface_exp}
In the last set of experiments, we evaluated \MN and baseline methods on the UTKFace dataset \cite{zhang2017age}. This dataset contains over $20,000$  face images with annotations of age, gender, and ethnicity. The age values range from $0$ to $116$, treated as a regression task. Gender is classified into binary categories, either male or female, while ethnicity is classified into five distinct categories, making it a multi-class classification task. We split the dataset according to $70\%-10\%-20\%$ to train, validation, and test datasets. Here, we use ResNet-18 for the shared network.

Results for this dataset based on $8$ random seeds are presented in \Tableref{tab:utkface}. Here as well \MN outperforms all methods, having the best results on $2$ out of $3$ tasks. Interestingly, our approach and MGDA, were the only methods to improve upon the STL baseline on the regression task. 

\section{Conclusions}
\label{sec:conclusions}
In this study, we present \MN, a novel method for aggregating the task gradients in MTL. Instead of treating the gradient of each task as a deterministic quantity we advocate here to assign a probability distribution over them. The randomness in them arises by noticing that there are many possible configurations for the task-specific parameters that work well. Hence, by tracking all of them using Bayesian tools we can obtain a richer description of the gradient space. This in turn allows us to model the uncertainty in the gradients and derive an update direction for the shared parameters that takes it into account. We demonstrate our method's effectiveness on several benchmark datasets compared with leading baseline methods. For future work, we would like to extend \MN beyond linear task heads. The challenge here would be to \emph{efficiently} estimate the Bayesian posterior and the gradient moments. Another possible limitation of \MN, having in common with other popular MTL methods, is that it may fail on rare or atypical examples \cite{sagawa2019distributionally}. 


\section*{Acknowledgements}
This study was funded by a grant to GC from the Israel Science Foundation (ISF 737/2018), and by an equipment grant to GC and Bar-Ilan University from the Israel Science Foundation (ISF 2332/18). IA is supported by a PhD fellowship from Bar-Ilan data science institute (BIU DSI). 

\section*{Impact Statement}
This paper presents work whose goal is to advance the field of Machine Learning. There are many potential societal consequences of our work, none which we feel must be specifically highlighted here.

\bibliography{References}
\bibliographystyle{icml2024}

\newpage
\appendix

\onecolumn
\section{Full Derivations}
\label{sec:full_derivations}
We now present the full derivation for \Eqref{eq:explicit_mom} \& \Eqref{eq:post_cls} presented in the main text. For clarity, we drop here the superscript notation of the task.

\subsection{Regression Moments}
\label{sec:full_derivations_reg}

Starting with the first moment,
\begin{equation}
\label{eq:first_moment}
\begin{aligned} 
    \E[\rvg_i] &= \int \rvg_i p(\rvg_i)d\rvg_i =  \int \rvg_i(\rvw) p(\rvw|\gD)d\rvw\\
    &= 2 \int \rvw (\rvh_i^T \rvw - y_i) p(\rvw|\gD)d\rvw\\
    &= 2 \int \rvw \rvw^T \rvh_i  - y_i \rvw p(\rvw|\gD)d\rvw\\
    &= 2 ([\rmS + \rvm \rvm^T]\rvh_i  - y_i \rvm).
\end{aligned}
\end{equation}
Where we made explicit the dependence in $\rvw$  on the first step. For computing the second moment we aided by the matrix reference manual \cite{GaussMoments},

\begin{equation}
\label{eq:second_moment}
\begin{aligned} 
    \E[\rvg_i \rvg_i^T] &= \int \rvg_i \rvg_i^T p(\rvg_i)d\rvg_i =  \int \rvg_i(\rvw) \rvg_i^T(\rvw) p(\rvw|\gD)d\rvw\\
    &= 4 \int \rvw (\rvh_i^T \rvw - y_i) (\rvh_i^T \rvw - y_i) \rvw^T p(\rvw|\gD)d\rvw\\
    &= 4 \int (y_i^2 \rvw \rvw^T \rvh_i  - 2 y_i \rvw \rvh_i^T \rvw \rvw^T + \rvw \rvw^T \rvh_i \rvh_i^T \rvw \rvw^T)p(\rvw|\gD)d\rvw.\\
    \end{aligned}
\end{equation}

We now solve each term separately and obtain the result,
\begin{equation}
\label{eq:second_moment_sep}
\begin{aligned} 
    \int \rvw \rvw^T p(\rvw|\gD)d\rvw &= \rmS + \rvm \rvm^T,\\
    \int \rvw \rvh_i^T \rvw \rvw^T p(\rvw|\gD)d\rvw &= \rvm \rvh_i^T(\rmS + \rvm \rvm^T) + (\rmS + \rvm \rvm^T)\rvh_i\rvm^T + \rvh_i^T\rvm(\rmS - \rvm \rvm^T),\\
    \int \rvw \rvw^T \rvh_i \rvh_i^T \rvw \rvw^T
    p(\rvw|\gD)d\rvw &= (\rmS + \rvm \rvm^T)(\rmA_i + \rmA_i^T)(\rmS + \rvm \rvm^T) + \rvm^T \rmA_i \rvm (\rmS - \rvm \rvm^T) + Tr(\rmA_i \rmS)(\rmS + \rvm \rvm^T).
    \end{aligned}
\end{equation}
Where, $\rmA_i = \rvh_i\rvh_i^T$.

\subsection{Second Order Posterior Approximation}
\label{sec:full_derivations_cls}

We now present the quadratic form of the log-posterior in \Eqref{eq:post_cls}. First we recap some of our notations here,
\begin{equation}
\label{eq:constants}
\begin{aligned} 
    c = log~p(\hat{\rvw} | \gB); \quad \rva = \left.- \frac{\partial log~p(\rvy | \rmX, \rvw)}{\partial \rvw} -  \frac{\partial log~p(\rvw)}{\partial \rvw} \right|_{\rvw=\hat{\rvw}}; \quad \rmB = \left.- \frac{\partial^2 log~p(\rvy | \rmX, \rvw)}{\partial \rvw^2}
    - \frac{\partial^2 log~p(\rvw)}{\partial \rvw^2}\right|_{\rvw=\hat{\rvw}}. 
    \end{aligned}
\end{equation}

Using these constants in \Eqref{eq:taylor_exp} yields the following form:
\begin{equation}
\label{eq:quad_form}
\begin{aligned} 
    &c + \rva^T (\rvw - \hat{\rvw}) + \frac{1}{2} (\rvw - \hat{\rvw})^T \rmB (\rvw - \hat{\rvw})\\
    =& \underbrace{c - \rva^T \hat{\rvw} + \frac{1}{2}(\hat{\rvw}^T\rmB\hat{\rvw}) - (\rmB^T\hat{\rvw} - \rva)^T\rmB^{-1}(\rmB^T\hat{\rvw} - \rva)}_{const.} + \frac{1}{2}(\rvw - (\hat{\rvw} - \rmB^{-1}\rva))^T\rmB(\rvw - (\hat{\rvw} - \rmB^{-1}\rva)).
    \end{aligned}
\end{equation}
The above takes the quadratic form of a Gaussian having mean $ (\hat{\rvw} - \rmB^{-1}\rva)$ and covariance $\rmB^{-1}$.

\begin{figure}[!t]
\centering
    \begin{subfigure}[Gender]{
    \includegraphics[width=0.4\linewidth]{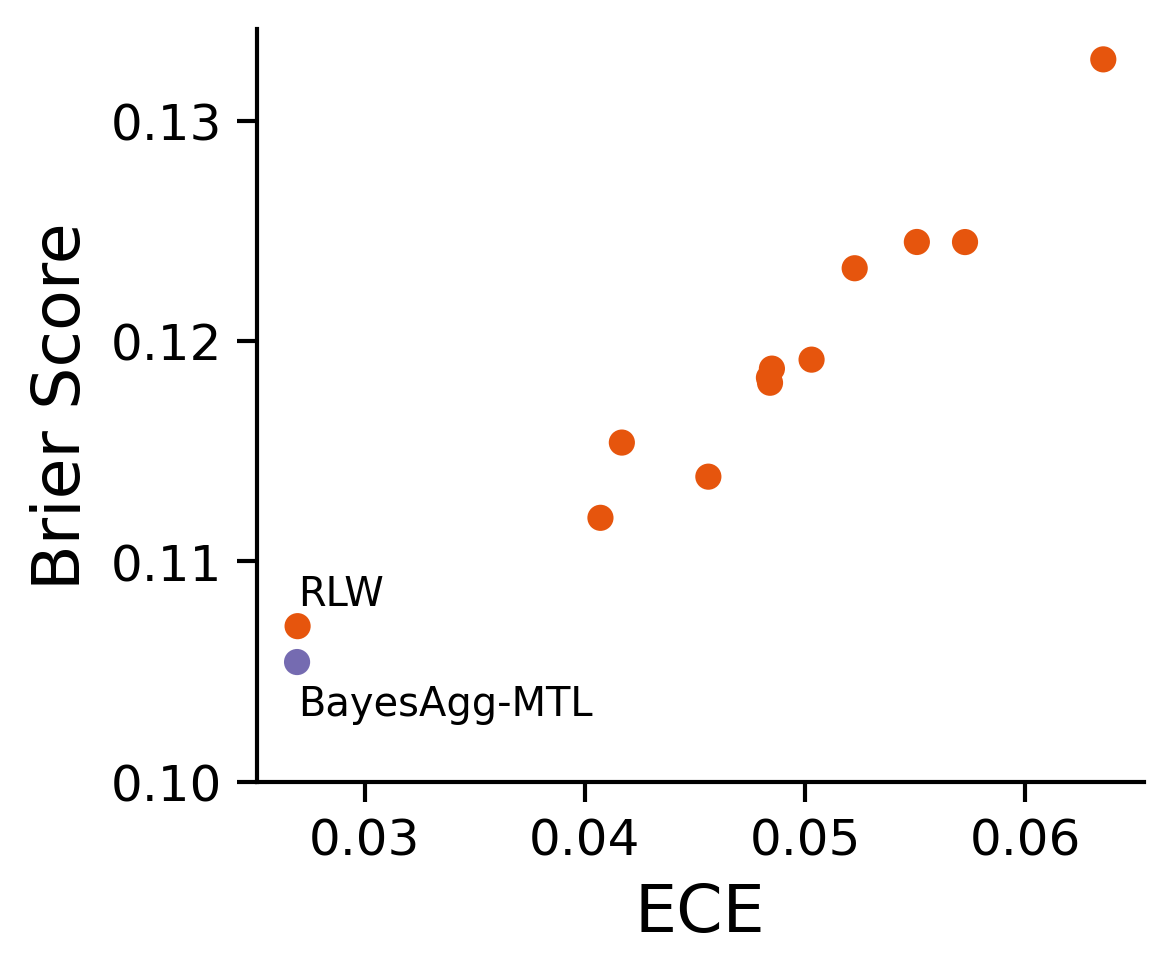}
    }
    \end{subfigure}
    \begin{subfigure}[Ethnicity]{
    \includegraphics[width=0.4\linewidth]{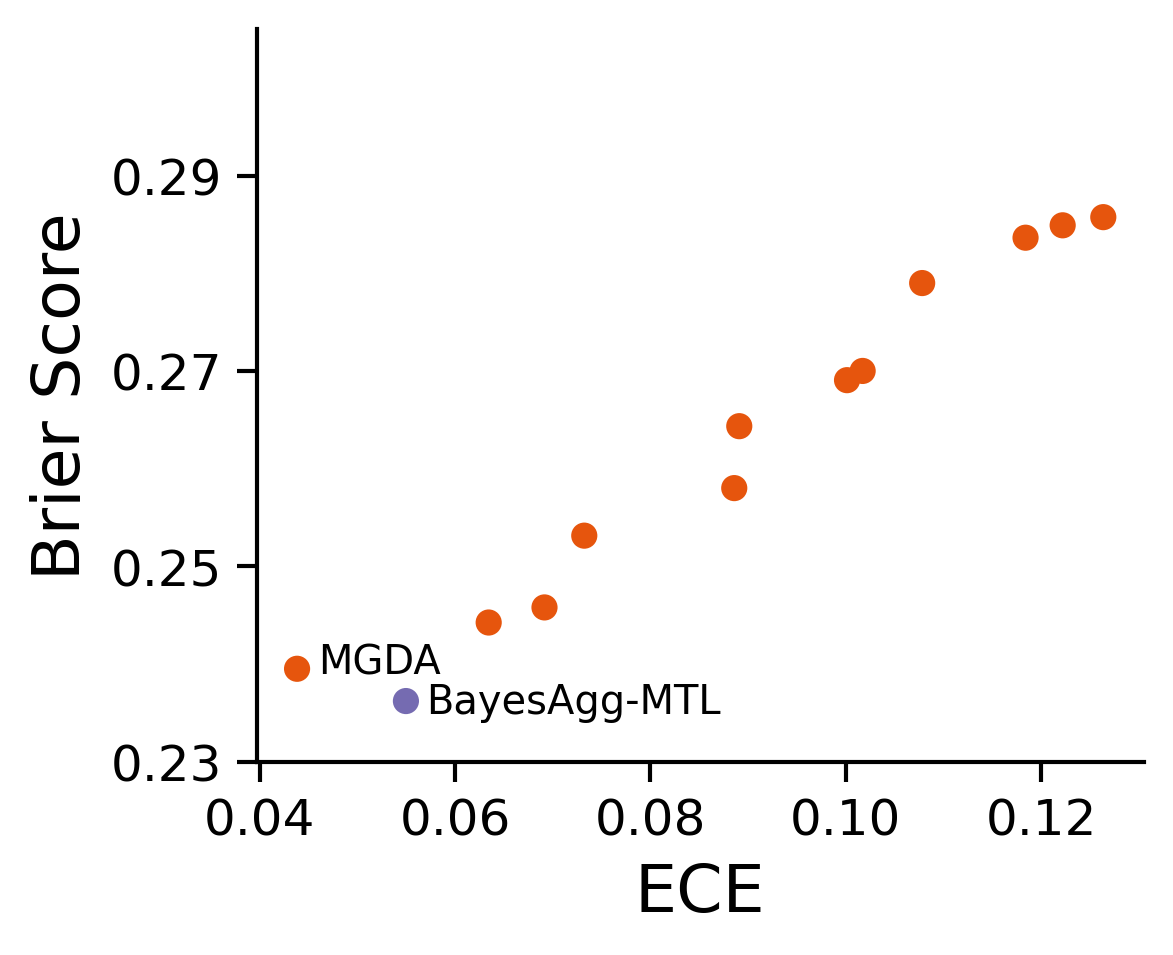}
    }
     \end{subfigure}
    \caption{Expected calibration error (ECE) vs Brier score for the Gender and Ethnicity tasks from the UTKFace dataset. In orange - baseline methods, and in purple our method. Lower values are better. We named our method and the top competitor on each plot.}
    \label{fig:calibration}
\end{figure}

\section{Full Experimental Details}
\label{sec:full_exp_details}
All the experiments were done using PyTorch on NVIDIA V100 and A100 GPUs having 32GB of memory. 

\textbf{QM9.} On this dataset we followed the training protocol presented by \citet{navon2022multi}. Specifically, We allocated $110,000$ examples for training and $10,000$ examples for validation and testing. The task labels are normalized to have zero mean and unit std. We use the implementation of \cite{fey2019fast} for the message-passing NN presented in \cite{gilmer2017neural} as the base NN. Here, we trained only our method and the baseline method Aligned-MTL-UB. All the other results were taken from \cite{navon2022multi, dai2023improvable}. We used the same random seeds as in those studies. Each method was trained for $300$ epochs using the ADAM optimizer \cite{Kingma2014AdamAM} with an initial lr of $1e-3$. The batch size was set to $120$. We use the ReduceOnPlate scheduler with the $\Delta_m$ metric, computed on the validation set. This metric was also used for early stopping and model selection. For \MN we set the number of pre-training epochs using linear scalarization to $50$. In initial experiments, we found that in regression tasks relatively higher values for the $s$ hyper-parameter were preferred. Hence, we searched over $s \in \{0.75, 0.85, 0.95\}$. For the Aligned-MTL-UB we did a hyper-parameter search over the scale modes in \{min, median, and rmse\}, and whether to apply that scale to the task-specific parameters as well.

\textbf{CIFAR-MTL.} Similarly to \cite{rosenbaum2018routing}, to form an MTL benchmark we used the coarse labels of CIFAR-100. Each example in the CIFAR-100 dataset belongs to one of $20$ super-classes. We use these super-classes as separate binary MTL tasks, where the task value is $1$ if the example indeed belongs to the super-class and $0$ otherwise. We use the official CIFAR train-test split of $50,000$ and $10,000$ respectively. We allocated $5,000$ examples from the training set to validation. To train the models we use a CNN having $3$ convolution layers with $160$ channels and a kernel of size $3$. Each convolution was followed by an Exponential Linear Unit (ELU) activation and max-pooling of $3 \times 3$. The final layer is a batch normalization layer. All methods were trained for $50$ epochs using the ADAM optimizer, with an initial learning rate of $1e-3$ and a scheduler that drops the learning rate by a factor of $0.1$ at $60\%$ and $80\%$ of the training. We set the batch size to $128$ and used a weight decay of $1e-4$. 
For all baseline methods, we did a hyper-parameter grid search over the most important $2-3$ hyper-parameters. Specifically, we would like to highlight that we searched over additional weight decay values for the LS, SI, and RLW baselines as advocated by \citet{kurin2022defense}.
As for \MN, unlike the regression case, for classification smaller $s$ values are preferred. We searched over $s \in \{5e^{-2}, 5e^{-3}, 5e^{-4}\}$. Also, we search over the number of pre-train epochs in $\{1, 3\}$. We set $J$, the number of Monte-Carlo samples, to $1024$, although we could have used much less without performance degradation. We used the validation accuracy for early stopping and model selection.

\textbf{ChestX-ray14.} This dataset reports the absence or appearance of $14$ types of chest diseases, which we view as an MTL problem. It contains approximately $112,000$ images from $32,717$ patients. We use the official data split presented in \cite{wang2017chestx}, having $70\%$ training examples, $10\%$ validation examples, and $20\%$ test examples. We follow the experimental setup of \cite{taslimi2022swinchex} that uses PyTorch Image Models \cite{rw2019timm} for data augmentations, a publicly available repository. We resize each image to size $224 \times 224$ and use data augmentation such as color jitter having $0.4$ intensity and random erase of pixels with a probability of $0.25$. Images are normalized according to ImageNet statistics \cite{russakovsky2015imagenet}. We use here ResNet-34 pre-trained on ImageNet as the shared feature extractor. We replaced the final classification layer with a fully connected layer of dimension $256$ followed by an ELU activation. Experimental details and hyper-parameter searches are similar to those described for CIFAR-MTL, except for the following changes. Here we trained for $100$ epochs, the batch size was set to $256$, and we didn't use a weight decay. We use the $\Delta_m$ metric for early stopping and model selection.

\textbf{UTKFace.} This dataset contains approximately $23,700$ images of faces, each associated with the age, gender, and ethnicity of the person. We remove $3$ examples from the dataset that have missing labels. We split the dataset to train/validation/test according to the $70-10-20$ scheme. The split was stratified by the age variable as it is the most diverse label. We treat the task of predicting the age as a regression task, and we normalize it to have zero mean and unit std. During training, images are resized to $140 \times 140$, randomly cropped to size $128$, and undergo random horizontal flip. Test images are resized and centered cropped. Here, we used ResNet-18 with the final classification layer replaced by a fully connected layer of size $256$ and an ELU activation. The experimental setup is similar to that described under the CIFAR-MTL, with the exception that here we trained for $100$ epochs. We perform a hyper-parameter grid search for all methods on this dataset as well. For our method, we set the number of pre-training epochs to $10$ and searched over $s \in \{0.3, 0.5, 0.8\}$ for the regression task and $s \in \{0.005, 0.05, 0.1\}$ for the classification tasks. We use the $\Delta_m$ metric for early stopping and model selection. Optimizing and evaluating the regression task is done using the MSE loss and the classification tasks using the standard cross-entropy loss.



\begin{table}[!t]
\small
\centering
\caption{Average run time (Sec. $\times 10^2$) of a training iteration on CIFAR-MTL and QM9 datasets.}
\vskip 0.11in
\begin{tabular}{@{}lcc@{}}
\toprule
 &  \textbf{CIFAR-MTL} & \textbf{QM9}\\ 
 \midrule
\multicolumn{1}{c}{LS}  & $1.551$ & $3.932$ \\
\multicolumn{1}{c}{SI}  & $1.600$ & $3.957$ \\
\multicolumn{1}{c}{RLW} & $1.555$ & $3.910$ \\
\multicolumn{1}{c}{DWA} & $1.571$ & $3.937$ \\
\multicolumn{1}{c}{UW}  & $1.740$ & $4.033$\\
\multicolumn{1}{c}{MGDA} & $15.34$ & $39.57$\\
\multicolumn{1}{c}{PCGrad} & $19.99$ & $29.89$ \\
\multicolumn{1}{c}{CAGrad} & $12.19$ & $26.16$ \\ 
\multicolumn{1}{c}{IMTL-G} & $10.12$ & $27.19$ \\ 
\multicolumn{1}{c}{Nash-MTL} & $22.47$ & $45.32$ \\
\multicolumn{1}{c}{IGBv2} & $3.651$ & $4.723$ \\
\midrule
\multicolumn{1}{l}{\textbf{Gradient w.r.t Representation}}\\
\midrule
\multicolumn{1}{c}{MGDA-UB} & $5.522$ & $9.352$\\
\multicolumn{1}{c}{IMTL-G} & $2.969$ & $4.426$ \\ 
\multicolumn{1}{c}{Aligned-MTL-UB} & $2.988$ & $4.428$ \\  %
\multicolumn{1}{c}{\MN (Ours)} & $5.558$ & $4.177$  \\
\bottomrule
\end{tabular}
\label{tab:run_time}
\end{table}

\section{Additional Experiments}
\subsection{Calibration}
\label{sec:calibration}
A possible benefit of using a Bayesian layer as the last layer is enhanced uncertainty estimation capabilities. Here we compare \MN to baseline methods on that aspect. To do so we log the expected calibration error (ECE) \cite{naeini2015obtaining} and Brier score \cite{brier1950verification} for all methods on the classification tasks of the UTKFace dataset. In ECE we first discretize the $[0, 1]$ line segment and then measure a weighted average difference between the classifier confidence and accuracy. We use $15$ interval bins in our comparison. Brier score measures the mean square error between the one-hot label encoding and the prediction probability vector. In both metrics, lower values are better. Results are presented in \Figref{fig:calibration}. From the figure, \MN is better calibrated than most methods on both datasets. On the gender task, it is best calibrated according to the two metrics. On the Ethnicity task, it has the best Brier score and second-best ECE score. We stress here that for a fair comparison with baseline methods, we did not use the Bayesian posterior of \MN on the last layer to make test predictions, but rather the point estimate of it learned during training. Using the full posterior should yield even better results.

\subsection{Training Time}
\label{sec:train_time}
\Tableref{tab:run_time} compares the run time of all methods on the CIFAR-MTL and QM9 datasets. We report the average processing time of a batch based on $10$ epochs.
To do the comparison, we use the best hyper-parameter configuration (in terms of performance) according to the CIFAR-MTL experiments. For MGDA and IMTL-G we present the run time under two settings, (1) when using in the aggregation scheme the full gradients w.r.t the shared parameters (top block); (2) when using in the aggregation scheme the gradients w.r.t the hidden layer (bottom block) as \MN does. For \MN we do not include the pre-training steps in the time measurements. From the table, methods that do not rely on the gradients for weighing the tasks are faster as outlined before in previous studies \cite{xin2022current, kurin2022defense}; however, this often comes at a significant performance reduction. \MN training time is almost as fast as those methods on regression problems, in which everything is done in closed-form, and slower on classification problems, partly due to the sampling process. Nevertheless, it is substantially faster than other gradient balancing methods that use gradients w.r.t the shared parameters.

\subsection{Comparison to Bayesian Training}
\label{sec:bayesian_train_comparison}

\begin{table}[!h]
\small
\centering
\caption{Comparison of $\Delta_m\%$ values to Deep Ensembles averaged over 3 random seeds.}
\vskip 0.11in
\begin{tabular}{@{}lcc@{}}
\toprule
 &  \textbf{QM9} & \textbf{UTKFace}\\ 
 \midrule
\multicolumn{1}{c}{Ensemble ($1024$ heads)}  & $~~161.4 \pm 13.1$ & $~0.99 \pm 0.62$ \\
\multicolumn{1}{c}{Ensemble ($10$ networks)}  & $~144.5 \pm 0.3$ & $-0.13 \pm 0.39$ \\
\midrule
\multicolumn{1}{c}{\MN (Ours)} & $\mathbf{~~53.2 \pm 7.1}$ &$\mathbf{-2.23 \pm 0.76}$\\
\bottomrule
\end{tabular}
\label{tab:standard_training}
\end{table}

Given that we used a Bayesian inference procedure in our approach, a natural question one may ask is \textit{how does standard approximate Bayesian training perform in MTL?} 

Recall that the goal of this paper is to use Bayesian inference on the last layer as a means to train deterministic MTL models using the uncertainty estimates in the gradients of the tasks. We use these uncertainty estimates to come up with an aggregation rule for combining the gradients of the tasks to a shared update direction. More concisely, our aim is to \textit{better learn a deterministic MTL model} while reducing as much as possible the computational overhead involved in training it. In standard approximate Bayesian training the gradient used in the backward process is considered as a deterministic quantity, similarly to non-Bayesian training. Hence, even when applying standard Bayesian inference to the task-specific parameters, the optimization issues regarding \textit{how to combine} the gradients of the tasks effectively remain.

To showcase that we compare \MN to deep ensembles \cite{lakshminarayanan2017simple} that have a strong link to approximate Bayesian methods \cite{wilson2020bayesian, d2021repulsive, wild2024rigorous}. We chose deep ensembles because of their simplicity and predictive abilities. We show here results on QM9 and UTKFace for two baselines: (1) Using $1024$ heads for each task and a shared backbone; (2) Using $10$ networks, each with a different backbone and task heads. The latter is substantially computationally more demanding as it requires different copies of the backbone as well, which is usually large. We combine the tasks using linear scalarization (i.e., equal weighting of the tasks) and averaging over the ensemble members. We follow the same experimental protocol of the paper and report the $\Delta_m$ values for each method in \Tableref{tab:standard_training}. From the table, the ensemble model with the shared backbone performs roughly the same as standard linear scalarization, with a slight advantage on QM9. This result makes sense as the uncertainty information is not taken into account when aggregating the gradients (i.e., only the mean values are used). Full ensemble training improves upon the ensemble baseline having a shared feature extractor, but it comes with a substantial computational overhead. Finally, \MN substantially outperforms both methods on both datasets.

\subsection{Full Results}
\label{sec:full_results}
In Tables \ref{tab:qm9_full} and \ref{tab:chest_full} we present the per-task results for all methods on the QM9 and ChestX-ray14 respectively. On QM9 we report the mean-absolute error of each task and on ChestX-ray14 the AUC-ROC of the tasks. Due to lack of space, we abbreviated several diseases names from the ChestX-ray14. We outline here the full names of all diseases: Atelectasis, Cardiomegaly, Consolidation, Edema, Effusion, Emphysema, Fibrosis, Hernia, Infiltration, Mass, Nodule, Pleural\_Thickening, Pneumonia, Pneumothorax.

\begin{table*}[!h]
\tiny
\centering
\caption{\textit{QM9}. Full test results averaged over 3 random seeds.}
    \vskip 0.11in
\begin{tabular}{@{}clcccccccccccc@{}}
\toprule
 &  & $\mu$ & $\alpha$ & $\epsilon_{\text{HOMO}}$ & $\epsilon_{\text{LUMO}}$ & $\langle R^2 \rangle$ & ZPVE & $U_0$ & $U$ & $H$ & $G$ & $c_v$ &  \\ \cmidrule(lr){3-13}
 &  & \multicolumn{11}{c}{MAE $\downarrow$} & $\mathbf{\Delta_m\%}$ ($\downarrow$) \\ 
 \midrule
\multicolumn{2}{c}{STL}&$0.067$&$0.181$&$60.57$&$53.91$&$0.503$&$4.53$&$58.80$&$64.20$&$63.80$&$66.20$&$0.072$&  \\ \midrule
\multicolumn{2}{c}{LS}&$0.106$&$0.326$&$73.57$&$89.67$&$5.197$&$14.06$&$143.4$&$144.2$&$144.6$&$140.3$&$0.129$& $177.6$ \\
\multicolumn{2}{c}{SI} &$0.309$&$0.346$&$149.8$&$135.7$&$1.003$&$4.51$&$55.32$&$55.75$&$55.82$&$55.27$&$0.112$& $77.8$ \\
\multicolumn{2}{c}{RLW}&$0.113$&$0.340$&$76.95$&$92.76$&$5.869$&$15.47$&$156.3$&$157.1$&$157.6$&$153.0$&$0.137$& $203.8$ \\
\multicolumn{2}{c}{DWA} &$0.107$&$0.325$&$74.06$&$90.61$&$5.091$&$13.99$&$142.3$&$143.0$&$143.4$&$139.3$&$0.125$& $175.3$ \\
\multicolumn{2}{c}{UW}&$0.387$&$0.425$&$166.2$&$155.8$&$1.065$&$5.00$&$66.42$&$66.78$&$66.80$&$66.24$&$0.123$& $108.0$ \\
\multicolumn{2}{c}{MGDA}&$0.217$&$0.368$&$126.8$&$104.6$&$3.227$&$5.69$&$88.37$&$89.41$&$89.32$&$88.01$&$0.120$& $120.5$ \\
\multicolumn{2}{c}{PCGrad}&$0.106$&$0.293$&$75.85$&$88.33$&$3.940$&$9.15$&$116.4$&$116.8$&$117.2$&$114.5$&$0.110$& $125.7$ \\
\multicolumn{2}{c}{CAGrad}&$0.118$&$0.321$&$83.51$&$94.81$&$3.219$&$6.93$&$114.0$&$114.3$&$114.5$&$112.3$&$0.116$& $112.8$ \\ 
\multicolumn{2}{c}{IMTL-G}&$0.136$&$0.288$&$98.31$&$93.96$&$1.753$&$5.70$&$101.4$&$102.4$&$102.0$&$100.1$&$0.097$& $77.2$ \\
\multicolumn{2}{c}{Nash-MTL}&$0.103$&$0.249$&$82.95$&$81.89$&$2.426$&$5.38$&$74.52$&$75.02$&$75.10$&$74.16$&$0.093$& $62.0$ \\
\multicolumn{2}{c}{IGBv2}&$0.251$&$0.333$&$149.1$&$130.2$&$0.956$&$4.39$&$56.75$&$57.19$&$57.25$&$56.73$&$0.110$& $67.7$ \\
\multicolumn{2}{c}{Aligned-MTL-UB}&$0.172$&$0.350$&$117.3$&$109.0$&$1.520$&$5.23$&$76.13$&$76.58$&$76.62$&$75.71$&$0.980$& $71.0$ \\
\midrule
\multicolumn{2}{c}{\MN (Ours)}&$0.122$&$0.280$&$87.78$&$90.44$&$1.776$&$5.31$&$63.33$&$64.91$&$66.71$&$81.91$&$0.093$& $\mathbf{53.2}$ \\
\bottomrule
\end{tabular}
\label{tab:qm9_full}
\end{table*}

\begin{table*}[!h]
\tiny
\centering
\caption{\textit{ChestX-ray14}. Full test results averaged over 3 random seeds.}
    \vskip 0.11in
\begin{tabular}{@{}clccccccccccccccc@{}}
\toprule
 &  & Atel. & Card. & Cons. & Edema & Effusion & Emphysema & Fibrosis & Hernia & Infi. & Mass & Nodule & Pleu. & Pneumonia & Pneu. \\ 
 \cmidrule(lr){3-16}
 &  & \multicolumn{14}{c}{AUC-ROC $\uparrow$} & $\mathbf{\Delta_m\%}$ ($\downarrow$) \\ 
 \midrule
\multicolumn{2}{c}{STL} & $.7543$ & $.8615$ & $.7132$ & $.8212$ & $.8224$ & $.6333$ & $.7357$ & $.7647$ & $.6830$ & $.6208$ & $.5894$ & $.6389$ & $.5710$& $.7701$ &  \\ \midrule
\multicolumn{2}{c}{LS}&$.7744$&$.8804$&$.7477$&$.8457$&$.8273$&$.8798$&$.8250$&$.9129$&$.7013$&$.8209$&$.7593$&$.7660$&$.7235$&$.8525$&$-14.62$\\
\multicolumn{2}{c}{SI}&$.7457$&$.8739$&$.7289$&$.8426$&$.8152$&$.8593$&$.7903$&$.8045$&$.6996$&$.7971$&$.7268$&$.7353$&$.6993$&$.8389$&$-1.94$\\
\multicolumn{2}{c}{RLW} &$.7596$&$.8704$&$.7389$&$.8385$&$.8218$&$.8390$&$.7956$&$.8646$&$.6991$&$.7933$&$.7340$&$.7362$&$.7101$&$.8345$&$-11.69$\\
\multicolumn{2}{c}{DWA} &$.7734$&$.8847$&$.7503$&$.8482$&$.8267$&$.8768$&$.8185$&$.9410$&$.6977$&$.8175$&$.7590$&$.7739$&$.7240$&$.8440$&$-14.79$\\
\multicolumn{2}{c}{UW} &$.7600$&$.8870$&$.7437$&$.8464$&$.8221$&$.8768$&$.8176$&$.9434$&$.7012$&$.8049$&$.7426$&$.7608$&$.7057$&$.8498$&$-13.95$\\
\multicolumn{2}{c}{MGDA} &$.7720$&$.8857$&$.7473$&$.8454$&$.8260$&$.8762$&$.8181$&$.9290$&$.6961$&$.8141$&$.7570$&$.7661$&$.7213$&$.8479$&$-14.44$\\
\multicolumn{2}{c}{PCGrad} &$.7678$&$.8793$&$.7461$&$.8432$&$.8266$&$.8721$&$.8165$&$.8565$&$.6991$&$.8123$&$.7499$&$.7629$&$.7203$&$.8451$&$-13.43$\\
\multicolumn{2}{c}{CAGrad} &$.7744$&$.8823$&$.7489$&$.8464$&$.8269$&$.8756$&$.8199$&$.9201$&$.6998$&$.8158$&$.7567$&$.7702$&$.7207$&$.8482$&$-14.50$\\
\multicolumn{2}{c}{IMTL-G} 
&$.7395$ &$.8533$ &$.7229$ &$.8235$ &$.8023$ &$.7692$ &$.7538$ &$.8973$ &$.6903$ &$.7543$ &$.7052$ &$.7221$ &$.6758$ &$.8026$ &$-8.24$\\
\multicolumn{2}{c}{Nash-MTL} &$.7623$&$.8774$&$.7445$&$.8420$&$.8206$&$.8627$&$.8214$&$.8997$&$.6999$&$.8035$&$.7412$&$.7553$&$.7117$&$.8447$&$-13.24$\\
\multicolumn{2}{c}{IGBv2} &$.7189$&$.8354$&$.7049$&$.8097$&$.7865$&$.7360$&$.7160$&$.7053$&$.6858$&$.6828$&$.6647$&$.7038$&$.6512$&$.7783$&$-2.83$\\
\multicolumn{2}{c}{Aligned-MTL-UB} &$.7689$&$.8801$&$.7491$&$.8456$&$.8245$&$.8772$&$.8221$&$.8992$&$.6997$&$.8115$&$.7543$&$.7674$&$.7208$&$.8497$&$-14.15$\\
\midrule
\multicolumn{2}{c}{\MN (Ours)} &$.7761$&$.8836$&$.7511$&$.8487$&$.8293$&$.8863$&$.8289$&$.9121$&$.6967$&$.8220$&$.7622$&$.7762$&$.7214$&$.8545$&$\mathbf{-14.96}$\\
\bottomrule
\end{tabular}
\label{tab:chest_full}
\end{table*}

\label{sec:additional_exp}


\end{document}